**Efficacy of MRI data harmonization in the age of machine learning: a multicenter study across 36 datasets**

Chiara Marzi[1,2], Marco Giannelli[3], Andrea Barucci[2], Carlo Tessa[4], Mario Mascalchi[5,6], Stefano Diciotti[7,8]

[1]Department of Statistics, Computer Science and Applications "Giuseppe Parenti", University of Florence, Florence, Italy
[2]"Nello Carrara" Institute of Applied Physics (IFAC), National Research Council (CNR), 50019 Sesto Fiorentino (Florence), Italy
[3]Unit of Medical Physics, Pisa University Hospital "Azienda Ospedaliero-Universitaria Pisana", 56126 Pisa, Italy
[4]Radiology Unit Apuane e Lunigiana, Azienda USL Toscana Nord Ovest, 54100 Massa, Italy
[5]Department of Experimental and Clinical Biomedical Sciences "Mario Serio", University of Florence, 50139 Florence, Italy
[6]Division of Epidemiology and Clinical Governance, Institute for Study, Prevention and netwoRk in Oncology (ISPRO), 50139 Florence, Italy
[7]Department of Electrical, Electronic, and Information Engineering "Guglielmo Marconi" - DEI, University of Bologna, 47522 Cesena, Italy
[8]Alma Mater Research Institute for Human-Centered Artificial Intelligence, University of Bologna, 40121 Bologna, Italy

Corresponding Author: Prof. Stefano Diciotti, stefano.diciotti@unibo.it

**Abstract**

Pooling publicly-available MRI data from multiple sites allows to assemble extensive groups of subjects, increase statistical power, and promote data reuse with machine learning techniques. The harmonization of multicenter data is necessary to reduce the confounding effect associated with non-biological sources of variability in the data. However, when applied to the entire dataset before machine learning, the harmonization leads to data leakage, because information outside the training set may affect model building, and potentially falsely overestimate performance. We propose a 1) measurement of the efficacy of data harmonization; 2) *harmonizer* transformer, i.e., an implementation of the *ComBat* harmonization allowing its encapsulation among the preprocessing steps of a machine learning pipeline, avoiding data leakage. We tested these tools using brain $T_1$-weighted MRI data from 1740 healthy subjects acquired at 36 sites. After harmonization, the site effect was removed or reduced, and we showed the data leakage effect in predicting individual age from MRI data, highlighting that introducing the *harmonizer* transformer into a machine learning pipeline allows for avoiding data leakage.

## Introduction

In recent years there has been an increasing trend toward data sharing in neuroimaging research communities, leading to a rising number of public neuroimaging databases and collaborative multicenter initiatives[1–4]. Indeed, pooling MRI data from multiple sites provides an opportunity to assemble more extensive and diverse groups of subjects[2,3,5,6], increase statistical power[3,7–10], and study rare disorders and subtle effects[11,12]. However, a major drawback of combining neuroimaging data across sites is the introduction of confounding effects due to non-biological variability in the data, typically related to image acquisition hardware and protocol. Indeed, properties of MRI such as scanner field strength, radiofrequency coil type, gradients coil characteristics, hardware, image reconstruction algorithm, and non-standardized acquisition protocol parameters can introduce unwanted technical variability, also reflected in MRI-derived features[13–15].

The harmonization of multicenter data, defined as applying mathematical and statistical concepts to reduce unwanted site variability while maintaining the biological content, is, therefore, necessary to ensure the success of cooperative analyses. Currently, among the harmonization methods for tabular data available to the neuroimaging scientific community, *ComBat* is one of the most widely used[7,12,16–34]. The ComBat model was first introduced in gene expression analysis as a batch-effect correction tool to remove unwanted variation associated with the site and preserve biological associations in the data[35]. In general, ComBat applies to situations where multiple features of the same type are measured for each participant, i.e., expression levels for different genes or imaging-derived metrics from different voxels or anatomical regions. The success of ComBat and derivatives has been measured compared to other harmonization techniques[3,5,6] and through simulations of the site effect from single-center data[2]. Previous literature has primarily focused on assessing the maintenance of biological variability in harmonized data[2,3,5]. However, less effort has been put into quantitative measurements of the efficacy of harmonization in removing the unwanted site effect.

Moreover, the pooling of multicenter data and the consequent availability of large sample sizes paves the way for data reuse with machine and deep learning techniques[17,19,22,23,25]. In the case of multicenter data, harmonization is thus added to conventional data preprocessing steps, including, e.g., data cleaning and imputation, feature extraction, and reduction. Similar to other procedures, the harmonization parameters should be optimized on training data only and subsequently applied to test data. Indeed, this approach avoids data leakage, which happens when information from outside the training set is used to create the model, potentially leading to falsely overestimated performance. Crucially, this aspect has sometimes been overlooked in previous applications of ComBat by harmonizing the entire data sample before data splitting (training and test sets) used for training and testing machine or deep learning techniques[2,5,17,19,22,23,25,36–41].

To the best of our knowledge, the harmonization techniques for neuroimaging data have been applied without paying attention to avoid data leakage, and this effect has not been quantified. In addition, despite the Python package *neuroHarmonize*[2] and the R code provided by Radua and colleagues[3] include functions that estimate the harmonization model on the training data and apply it separately to the test data, they have not been conceived to be executed on a machine learning *pipeline*, i.e., an end-to-end framework that orchestrates the flow of data into a machine learning model and allows to speed-up the development and test of machine learning systems, natively avoiding data leakage.

For these reasons, in this study, we propose 1) a measurement of the efficacy of data harmonization in reducing the site effect by the performance of a machine learning classifier trained to identify the imaging site, 2) a ComBat implementation using a *harmonizer* transformer, i.e., a method that, combined with a classifier/regressor, forms a composite estimator, to be used in a machine learning pipeline, thus simplifying data analysis and avoiding data leakage (the source code of the efficacy measurement and *harmonizer* transformer are publicly available in a GitHub repository at https://github.com/Imaging-AI-for-Health-virtual-lab/harmonizer). First, we showed and measured the effect of data leakage when harmonization is performed before data splitting using simulated neuroimaging data with known site effect. Then, we estimated the efficacy of data harmonization in reducing the site effect using the *harmonizer* transformer on brain $T_1$-weighted MRI data from 1787 healthy subjects aged 5 – 87 years acquired at 36 imaging sites. The morphological features of cortical thickness (CT) and fractal dimension (FD), a descriptor of the structural complexity of objects with self-similarity properties[42], are extracted to characterize brain morphology. To the best of our knowledge, this is the first time that measures of brain structural complexity, such as FD, have been studied on such a large, multicenter, and harmonized data sample. Finally, we investigated the age prediction using neuroimaging variables harmonized in the entire dataset before machine learning and using the *harmonizer* transformer to estimate the effect of data leakage in *in vivo* data.

## Methods

### *MRI datasets*

We gathered brain MR $T_1$-weighted images of 1787 healthy subjects aged 5 – 87 years belonging to 36 single-center datasets of various studies. These include the *Autism Brain Imaging Data Exchange (ABIDE)* (http://fcon_1000.projects.nitrc.org/indi/abide/) first and second initiatives (ABIDE I and ABIDE II, respectively)[43,44], the *Information eXtraction from Images (IXI)* study (https://brain-development.org/ixi-dataset/), the *1000 Functional Connectomes Project (FCP)* (http://fcon_1000.projects.nitrc.org/fcpClassic/FcpTable.html), and the *Consortium for Reliability and Reproducibility (CoRR)* (http://fcon_1000.projects.nitrc.org/indi/CoRR/html/index.html). From each study, we drew several specific datasets of brain MR $T_1$-weighted images acquired in the same place with the same scanner and acquisition protocol (see Table 1). Both ABIDE I, and II initiatives contributed with 17 datasets, and we named them with the initiative prefix (ABIDEI or ABIDEII) followed by the institution name that collected the images (e.g., ABIDEI-CALTECH and ABIDEII-BNI_1). For the institution names, we used the same nomenclature as reported online[45] with the following exceptions: (i) we merged LEUVEN_1 and LEUVEN_2 data in ABIDEI-LEUVEN, UCLA_1, and UCLA_2 data in ABIDEI-UCLA, UM_1 and UM_2 data in ABIDEI-UM, because the acquisition parameters were the same, (ii) we split the data from ABIDEII-KKI_1 into ABIDEII-KKI_8ch and ABIDEII-KKI_32ch, because the acquisitions were performed using an 8-channel or a 32-channel phased-array head coil, respectively. The IXI study provided three different datasets named with the prefix IXI followed by the institution name that collected the images (e.g., IXI-Guys). From the 1000 FCP and CoRR studies, we used the *International Consortium for Brain Mapping (ICBM)* and the *Nathan Kline Institute - Rockland Sample Pediatric Multimodal Imaging Test-Retest Sample (NKI2)* datasets, respectively.

In each single-center dataset, baseline MRI scans of typically developing and aging brain (one for each subject) with available age and sex information were included. The lack of a recognized neurological or psychiatric disorder diagnosis was used to define normal development and aging. The leading institutions, at each site where the MR images were collected, had obtained informed consent from all participants, and were authorized by the local Ethics Committees. Table S1 shows the general characteristics of each single-center dataset. In this study, we grouped the single-center datasets into three multicenter meta-datasets based on age and the amount of overlap between age distributions. We have considered the following age ranges: childhood (5 – 13 years), adolescence (11 – 20 years), and adulthood (18 – 87 years). We measured the overlap between age distributions by the n-distribution Bhattacharyya coefficient (BC)[46], an extension of the 2-distribution BC[47]. The BC coefficient is 0 when there is no overlap and 1 when the overlap is complete. In our study, n is the number of the single-center datasets grouped in the meta-dataset covering the above-mentioned age ranges and may be different in every meta-dataset. Therefore, we constructed the CHILDHOOD meta-dataset containing 11 single-center datasets, whose subjects' age varies between 5 and 13 years, and age distributions have a BC = 0.71. The ADOLESCENCE meta-dataset includes 9 single-center datasets whose subjects' age ranges from 11 to 20 years, and age distributions have an overlap amount equal to 0.45. Finally, the ADULTHOOD meta-dataset consists of all data belonging to subjects aged between 18 and 87 years old (12 single-center datasets), whose age distributions have a BC = 0. A detailed description of the composition of each meta-dataset and their age distributions are shown in Table 2 and Figure 1, respectively. In addition, we also merged all single-center datasets, creating a meta-dataset, called LIFESPAN, that covers the entire age range (5 – 87 years). In this meta-dataset, composed of 36 imaging sites, the single-center age distributions have a null overlap (Figure 1).

***MR image processing***

For each brain MR $T_1$-weighted image, we performed a cortical reconstruction and a volumetric segmentation (see details in section 2.2.1). In this work, we analyzed cerebral structures only, and we extracted neuroimaging features from various regions of the cerebral cortex: the entire cerebral cortex, separately the left/right hemispheres of the cerebral cortex, and left/right frontal, temporal, parietal, and temporal lobes. In particular, for each region, we computed the average cortical thickness (CT) and the fractal dimension (FD) (see details in section 2.2.2).

*Cortical reconstruction and volumetric segmentation*

We used the FreeSurfer package to perform completely automated cortical reconstruction and volumetric segmentation of each subject's structural $T_1$-weighted scan. We used version 7.1.1, except in a few cases: (i) for $T_1$-weighted images belonging to ICBM and NKI2 datasets, we used FreeSurfer version 5.3, and (ii) for the ABIDEI datasets, we used the FreeSurfer version 5.1 outputs previously made available online by Cameron and colleagues[48] (http://preprocessed-connectomes-project.org/abide/index.html). Even though different FreeSurfer versions may affect neuroimaging variables[49–53], such variability is considered part of the site variability and handled by the harmonization procedure. Indeed, all subjects in each center have been processed with the same version of FreeSurfer. FreeSurfer is extensively documented (see Ref.[54] for a review) and publicly accessible (http://surfer.nmr.mgh.harvard.edu/). In addition to the standard FreeSurfer outputs, we performed a parcellation of the cortical lobes using the *mri_annotation2label* tool with the --*lobesStrict* option.

All Freesurfer outputs used in this study were visually inspected for quality assurance by two experienced radiologists (M.M. and C.T., with 35 and 30 years of experience, respectively) following an improved version of the ENIGMA Cortical Quality Control Protocol 2.0 (http://enigma.ini.usc.edu/protocols/imaging-protocols/). Firstly, we created an HTML file for each single-center dataset showing, for each subject, the segmentation of the cortical regions overlayed on the $T_1$-weighted images. Then, we scrolled the HTML file to determine gross segmentation errors in any cortical regions visually. For each single-center dataset, we estimated the statistical outliers for CT features (see section 2.2.2 for details about CT features extraction), defined as any data points below or above the mean by 2.698 standard deviations. For each subject, we carefully inspected the cortical segmentations that showed features values labeled as statistical outliers to assess whether the outlier was an actual segmentation error. In this case, the subject was excluded from further analyses.

*Extraction of cortical thickness and fractal dimension features*

For each subject, using FreeSurfer tools, we computed the average CT of each cortical region as the average distance measured from each vertex of the gray/white boundary surface to the pial surface[55].

The FD is a numerical representation of shape complexity[56]. The FD is normally a fractional value and is considered a dimension because it gives a measure of space-filling capacity[57]. An FD value between 2 and 3 is typical of a complex and heavily folded 2-D surface buried in a 3-D region, such as the human cerebral cortex. The FD is a very compact measure of shape complexity, combining cortical thickness, sulcal depth, and folding area into a single numeric value[58,59]. In this study, the fractal analysis was carried out using the *fractalbrain* toolkit version 1.0 (freely available at https://github.com/chiaramarzi/fractalbrain-toolkit) and described in detail in Marzi et al., 2020[59]. The *fractalbrain* toolkit processes FreeSurfer outputs directly, computing the FD of various regions of the cerebral cortex: the entire cerebral cortex, separately the left/right hemispheres, and left/right frontal, temporal, parietal, and temporal lobes. *Fractalbrain* performs the 3D box-counting algorithm[60], adopting an automated selection of the fractal scaling window[59] – a crucial step for establishing the FD for non-ideal fractals[59,61].

Briefly, we overlapped a grid composed of 3D cubes of different sizes $s$ (where $s = 2^k$ voxels, and k = 0, 1, ..., 8) onto the segmentation and recorded the number of cubes $N(s)$ needed to fully enclose the structure for each size. This process was repeated with 20 uniformly distributed random offsets to prevent the systematic influence of the grid placement, and the relative box count was averaged to obtain a single $N(s)$ value[62,63]. For a fractal object, the data points of the number of cubes $N(s)$ vs. size $s$ in the log-log plane can be modeled through a linear regression within a range of spatial scales called the fractal scaling window. *Fractalbrain* automatically selects the optimal fractal scaling window by searching for the interval of spatial scales that provides the best linear fit, as measured by the rounded coefficient of determination adjusted for the number of data points ($R^2_{adj}$). If multiple intervals have the same rounded $R^2_{adj}$, the widest interval (i.e., the one that contains the most data points in the log-log plot) is selected[59]. The FD of the brain structure is then estimated as the slope (in absolute value) of the linear regression model included in the automatically selected fractal scaling window. As an example, in Figure 2, we reported a log-log plot of the 3D box-counting algorithm optimized for the automatic selection of the best fractal scaling window of the cerebral cortex of one subject.

### *Harmonization of brain cortical features*

We harmonized cortical features using ComBat, a model that builds upon the statistical harmonization technique proposed by Johnson and colleagues[35] for location and scale (L/S) adjustments to the data while preserving between-subject biological variability. Briefly, let $y_{ijf}$ be the one-dimensional array of $n$ neuroimaging features for the single-center $i$, participant $j$, and feature $f$, for a total of $k$ single-center datasets, $n$ participants, and $V$ features. Still, let $X$ be the $n \times p$ matrix of biological covariates of interest, and Z be the $n \times k$ matrix of single-center labels. The ComBat harmonization model can be written as follows:

$$y_{ijf} = f_f(X_{ij}) + Z_{ij}\vartheta_f + \delta_{if}\varepsilon_{ijf} \tag{1}$$

where $f_f(X_{ij})$ denotes the variation of $y_{ijf}$ captured by the biologically relevant covariates $X_{ij}$, $\vartheta_f$ is the one-dimensional array of the $k$ coefficients associated with the single-center labels $Z_{ij}$ for the feature $f$. We assume that the residual terms $\varepsilon_{ijf}$ have mean 0. The parameters $\delta_{if}$ describe the multiplicative site effect of the *i-th* site on the feature $f$, i.e., the scale (S) adjustment, while the location (L) parameter for the *i-th* site on the feature *f*, is represented by $\gamma_{if}$ (the empirical Bayes estimates of the term $Z_{ij}\vartheta_f$). Consistent with the ComBat model notation used in Fortin et al. (2017), the harmonized $y^*_{ijf}$ become:

$$y^*_{ijf} = \frac{y_{ijf} - f_f(X_{ij}) - \gamma_{if}}{\delta_{if}} + f_f(X_{ij}) \qquad (2)$$

In this study, we used the ComBat model implemented in the *neuroHarmonize* v. 2.1.0 package (freely available at https://github.com/rpomponio/neuroHarmonize) – an open-source and easy-to-use Python module[2]. In particular, *neuroHarmonize* extends the *neuroCombat* package[5,6] with the possibility of specifying covariates with generic nonlinear effects on the neuroimaging feature to harmonize. In particular, the $f_f(X_{ij})$ term in Equation (1) is a Generalized Additive Model (GAM) function of the specific covariates[2]. Indeed, MRI-derived features are known to be influenced by demographic factors, such as age[2,3,5,59,64–70] and sex[71]. In our study, these variables were included in the harmonization process as sources of inter-subject biological variability. Finally, since it is not evident that the site effect affects all MRI-derived measures in the same way[3], we performed a separate harmonization for each feature group of the same type (i.e., CT and FD).

*The harmonizer transformer*

The increased sample size due to the pooling of data acquired in various centers necessarily facilitates the application of machine learning techniques. For training and testing machine learning models, a proper validation scheme that handles data splitting must be chosen (Figure 3). This choice is crucial to avoid data leakage by ensuring that the entire workflow (preprocessing and model-building steps) is constructed on training data and evaluated on test data never seen during the learning phase. Indeed, data leakage in the training process may incur falsely high performance in the test set (see, e.g., Ref.[72] and Ref.[73]). Especially in Medicine and Healthcare, where relatively small datasets are usually available, the straightforward hold-out validation scheme is rarely applied. In contrast, the cross-validation (CV) and its nested version (nested CV) for hyperparameters optimization of the entire workflow[74–76] are frequently preferred. Also, repeated CVs or repeated nested CVs are suggested for improving the reproducibility of the entire machine learning system[75]. Several training and test data procedures are carried out in all these validation schemes on different data split, recalling the need for a compact code structure to avoid errors that may lead to data leakage. In this view, machine learning pipelines are a solution because they orchestrate all the processing steps in a short, easier-to-read, and easier-to-maintain code structure (Figure 3). A pipeline represents the entire data workflow, combining all transformation steps (e.g., data cleaning, data imputation, data scaling, and general data preprocessing) and machine learning model training. It is essential to automate an end-to-end training/test process without any form of data leakage and improve reproducibility, ease of deployment, and code reuse, especially when complex validation schemes are needed.

In the Scikit-learn library, a popular, open-source, well-documented, and easy-to-learn machine learning package that implements a vast number of machine learning algorithms, a pipeline is a chain of “transformers” and a final “estimator” acting as a single object. The transformers are modules that apply preprocessing to the data, whereas estimators are modules that fit a model based on training data and are capable of inferring some properties on new data (https://scikit-learn.org/stable/developers/develop.html). In particular, transformers are classes with a “fit” method, which learns model parameters (e.g., mean and standard deviation for data standardization) from a training set, and a “transform” method which applies this transformation model to any data. For example, for data standardization (transforming data to have zero mean and unit standard deviation), the mean $\mu$ must be subtracted from the data, and the result must be divided by the standard deviation $\sigma$. Notwithstanding, this procedure must be firstly performed on the training set (using $\mu$ and $\sigma$ computed in the training set). In the test set, or any validation set, the same transformation must be applied to data using the same two parameters $\mu$ and $\sigma$ computed for centering the training set. Basically, the “fit” method calculates the parameters (e.g., $\mu$ and $\sigma$ in our case) and saves them internally, whereas the “transform” method applies the transformation (using the saved parameters) to any particular set of data.

For these reasons, in this study, we propose the *harmonizer* – a Scikit-learn Python transformer that encapsulates the *neuroHarmonize* procedure among the preprocessing steps of a machine learning pipeline. The “fit” method of the *harmonizer* transformer learns the *NeuroHarmonize* model parameters from a training set and saves the parameters internally, whereas the “transform’” method is used to apply the *neuroHarmonize* model, previously learned on the training data set, e.g., to unseen data. The source code of the *harmonizer* transformer is publicly available in a GitHub repository at https://github.com/Imaging-AI-for-Health-virtual-lab/harmonizer.

In the following, we included the *harmonizer* transformer in a pipeline to learn the harmonization procedure parameters on the training data only and apply the harmonization procedure (with parameters obtained in the training set) to the test data. This prevented data leakage in the harmonization procedure independently of the chosen validation scheme.

### *Statistical and machine learning analyses*

We performed the statistical and machine learning analyses described in the following paragraphs for each feature group of the same type (i.e., CT and FD) and each meta-dataset (i.e., CHILDHOOD, ADOLESCENCE, ADULTHOOD, and LIFESPAN).

#### *Visualization and quantification of site effect*

We first performed a series of analyses of increasing complexity to explore the actual existence of a site effect in the data. For each region-feature pair, we qualitatively showed the site effect on raw data through boxplots, using the site as the independent variable and each region-feature pair as the dependent variable. Quantitatively, the site effect was measured by analyzing covariance (ANCOVA) – a general linear model that blends analysis of variance (ANOVA) and linear regression. ANCOVA evaluates whether the means of a dependent variable are equal across levels of a categorical independent variable while statistically controlling for the effects of other variables that are not of primary interest, known as covariates or nuisance variables. In this study, we set the single-center dataset as the independent variable, age, $age^2$, and sex as covariates, and each region-feature pair as the dependent variable.

Additionally, to further investigate the site effect on raw data and to measure the success of ComBat harmonization, we predicted the imaging site from the neuroimaging features, grouped by feature type, namely CT and FD. Specifically, we used the supervised eXtreme Gradient Boosting (XGBoost) method (with version 0.90 default hyperparameters for a classification task), a scalable end-to-end tree-boosting system widely used to achieve state-of-the-art performance on many recent machine learning challenges[77]. Using N=100 repetitions of a stratified 5-fold CV, we estimated the median balanced accuracy. The statistical significance of prediction performance was determined via permutation analysis. Thus, for each features group, 5000 new models were created using a random permutation of the target labels (i.e., the imaging site), such that the explanatory neuroimaging variables were dissociated from their corresponding imaging site to simulate the null distribution of the performance measure against which the observed value was tested[78]. Since, in this study, single-center datasets showed different age groups, the random target labels permutation was performed within groups of subjects of similar age[79], which were categorized into five-year intervals. The selection of a 5-year value was made to ensure it was sufficiently small to discern age differences while being large enough to avoid an excessive reduction in the potential permutations within each age group.

Median balanced accuracy was considered significantly different from the chance level when the p-value computed using permutation tests was < 0.05. Additionally, we calculated the average confusion matrix over repetitions to graphically evaluate the goodness of prediction. The same imaging site prediction was performed on raw data (i.e., without harmonization) to confirm the existence of the site effect and on harmonized data (with *neuroHarmonize* and *Harmonizer* transformer) to investigate if the site effect was reduced or removed.

We propose to measure the efficacy of harmonization in reducing or removing the site effect through a two-step assessment. First, we evaluated whether the site prediction after the harmonization process was not significantly different from a random prediction by comparing the median balanced accuracy over repetitions with the distribution of balanced accuracies estimated using the permutation test with 5000 permutations (the default value in FSL – FMRIB Software library – *randomise* tool for non-parametric permutation inference on neuroimaging data[80]). Considering, for example, a significance threshold of 0.05 in the permutation test, in the case of complete removal of the site effect, the site prediction will not be different from that of a random model (i.e., p-value ≥ 0.05). Second, in the case of permutations test p-value < 0.05, we compared the balanced accuracy obtained by predicting the site without and with the harmonization procedure. In particular, we assessed the site effect reduction by ensuring that the median balanced accuracy obtained predicting the imaging site with harmonized data was significantly lower than that estimated with raw data through the non-parametric one-sided Wilcoxon signed-rank test, with a significance threshold of 0.05[81]. The source code for evaluating the effectiveness of harmonization using the *harmonizer* transformer is publicly available in a GitHub repository at https://github.com/Imaging-AI-for-Health-virtual-lab/harmonizer.

To estimate the effect of data leakage in the prediction of the imaging site caused by performing the harmonization on all data before splitting into training and test sets, we tested whether the balanced accuracies obtained using *neuroHarmonize* on all data before any split were consistently lower than those estimated using the *harmonizer* transformer in the above mentioned stratified CV scheme. Since the same data set splits were applied for both CT and FD, the comparison was carried out through a paired test, i.e., the non-parametric one-sided Wilcoxon signed-rank test with a significance threshold of 0.05[81].

*Associations with age*

While it is essential to show that a harmonization method successfully reduces a possible site effect, it is equally crucial to note that it preserves the biological variability in the data. Indeed, a harmonization method that removes both site and biological effects has no utility. One of the most influential sources of biological variability in the neuroimaging features of healthy subjects is undoubtedly chronological age. Throughout the lifespan, the brain structure changes because of a complex interplay between multiple maturational and neurodegenerative processes. Such processes could yield large spatial and temporal variations in the brain[65,82,83].

For these reasons, we attempted to predict individual age from neuroimaging features through an XGBoost model (version 0.90 with default hyperparameters for a regression task)[77]. We estimated the median (over repetitions) mean absolute error (MAE) using N=100 repetitions of a 5-fold CV. Age prediction was performed on harmonized data using both *neuroHarmonize* and the *harmonizer* transformer in the CV pipeline. To estimate the effect of data leakage in the age prediction caused by performing the harmonization on all data before splitting into training and test sets, we compared the MAE values obtained using *neuroHarmonize* on all data before any split and the *harmonizer* transformer in the above-mentioned CV scheme. In particular, since the same data set splits were applied for both CT and FD, we assessed whether the median MAE using *neuroHarmonize* on all data before any split was consistently lower than that estimated using the *harmonizer* transformer through a paired test, i.e., the non-parametric one-sided Wilcoxon signed-rank test with a significance threshold of 0.05[81].

Moreover, before and after the harmonization procedure, for each region-feature pair, we qualitatively visualized the site effect on the relationship between age and each region-feature pair through scatterplots (with age as the independent variable and each region-feature pair as the dependent variable).

***Simulation experiments***

The *harmonizer* transformer prevents data leakage in the harmonization procedure in any machine learning pipeline independently of the chosen validation scheme. Differently, applying harmonization before data spitting, data leakage is present, and its severity depends on the specific context and the extent of the leakage. In Neuroimaging, the entity and impact of the data leakage effect is still an underexplored area. Therefore, we performed simulation experiments (with known site effects) and computational tests for assessing the data leakage effect when the harmonization process is performed before the training-test data splitting.

*CT and FD data simulation settings*

Let $y_{ijf}$ be the one-dimensional array of the simulated feature $f$, for the single-center $i$, and participant $j$, for a total of $k$ single-center datasets, $n_i$ participants for each center, and $V$ features. In this study, we simulated CT and FD data for $\mathrm{k} = 3, 10, 36$ single-centers. Each single-center dataset provided the same number of participants (i.e., $n_i = \mathrm{n}$), with $n$ assuming the values $25, 50, 100, 250$. Totally, we did 24 experiments, i.e., we simulated 24 different multicenter datasets (12 for the CT features and 12 for the FD measures).

Each $y_{ijf}$ was generated based on the model proposed by Johnson and colleagues[35] and recently used for neuroimaging features' simulation by Chen and collaborators[84]:

$$y_{ijf} = \alpha_f + \beta_{f1} x_{ij} + \beta_{f2} x_{ij}^2 + \gamma_{if} + \delta_{if} \varepsilon_{ijf} \quad (3)$$

where $\alpha_f$ is the average value of the feature $f$ in the single-center ICBM dataset, $\beta_{f1} = -0.0009$ and $\beta_{f2} = -0.00005$ are the linear and quadratic effects of the age on the feature $f$, respectively, and $x_{ij}$ is a simulated age variable drawn from a uniform distribution $\mathrm{X} \sim \mathrm{uniform}([20,90])$. Considering the nature of our investigation, which examines the relationship between cortical thickness and FD with age, it is reasonable to assume that the relationship is no more than quadratic[59,85]. The mean site effect $\gamma_{if}$ was drawn from a normal distribution with zero mean and standard deviation equal to 0.1, while the variance site effect $\delta_{if}$ was drawn from a center-specific inverse gamma distribution with chosen parameters. For our simulations, we chose to distinguish the site-specific location factors by assuming independent and identically distributed (i.i.d.) normal distributions and scaling factors using the parameters described as follows. We set the value of the inverse gamma shape, for each center, as {46, 51, 56}, respectively, when k = 3, as {40, 42, .., 58} when k = 10, and as {10, 12, .., 40, 41, .., 50, 52, .., 70} when k = 36. In all cases, the inverse gamma scale was set to 50.

*Measuring the effect of data leakage*

We measured the effect of data leakage for both the site and age prediction independently. Hereinafter, we will refer generically to performance, indicating the balanced accuracy for the site prediction task and the MAE for the age prediction task. To measure the effect of data leakage, after an external hold-out (Figure 4), firstly, we computed the performance of an imaging site/age prediction estimator trained using a) the *harmonizer* transformer within the machine learning pipeline (internal not leaked test set) and b) harmonizing all data with *neuroHarmonize* before the actual prediction (internal leaked test set). Secondly, we compared these performances with that observed on an external test set never used for harmonization and training (Figure 4). In the absence of data leakage, the performance in the internal and external test sets should be similar and not significantly different. When data leakage is present, the performance in the internal test set is overly optimistic (i.e., significantly better than that on the external test set). In detail, for each experiment, we performed the following steps.

*External hold-out*. We randomly split the data into two parts, i.e., a *data set* containing 50% of the samples and an *external test set* with the other 50% of the instances.

*Imaging site/age prediction estimator training and test on the external test set*. We fitted a harmonization model with *neuroHarmonize* using age as a covariate with a nonlinear relationship with individual MRI-derived features. To fit the harmonization model, we used the same number of instances adopted for the other two approaches (see next analyses), i.e., 80% of samples, randomly chosen, of the data set. Then, we applied the harmonization model to the *data set* and the *external test set*. Finally, we trained an XGBoost model (with version 0.90 default hyperparameters for a classification task) to predict the imaging site/age and tested it on the harmonized external test set.

*Imaging site/age prediction estimator training and test using harmonizer transformer within the machine learning pipeline (not leaked internal test set)*. We trained and tested a pipeline containing the harmonizer transformer and an XGBoost estimator (with version 0.90 default hyperparameters for a classification task) on the *data set* to predict the imaging site/age through a stratified 10-times repeated 5-fold CV. Thus, we trained the pipeline in the training sets of each iteration of the CV and considered the performance within the test sets of the CV.

*Imaging site/age prediction estimator training and test harmonizing all data with neuroHarmonize before imaging site prediction (leaked internal test set)*. We trained and tested a pipeline containing an XGBoost estimator (with version 0.90 default hyperparameters for a classification task) on the harmonized dataset to predict the imaging site/age through a stratified 10-times repeated 5-fold CV. Thus, we trained the pipeline in the

training sets of each iteration of the CV and considered the performance metric within the test sets of the CV.

For each task, i.e., imaging site and age prediction, we repeated each experiment (i.e., all these steps) 100 times with random data splits and computed the average performance across the 100 repetitions. Finally, we compared the average performance across the 100 repetitions of each internal test set (leaked and not-leaked) with that of the external test set. When data leakage is present, the performance in the internal test set is better than that on the external test set (i.e., lower balanced accuracy and MAE values for the imaging site and age prediction, respectively). To assess whether the average performance of each internal test set was lower than that of the external test set, we conducted a one-tailed t-test, applying Bonferroni correction for multiple comparisons. This statistical analysis allowed us to evaluate the significance of any differences observed between the average performance of the internal and external test sets.

In addition, we calculated, for each internal test set, the Cohen's d effect size to estimate the magnitude of the differences between performance distributions' means. Specifically, we used the following Cohen's d formula: $\mathrm{d} = \frac{\overline{x_e} - \overline{x_\iota}}{s}$ where $\overline{x_e}$ is the average performance in the external test set, $\overline{x_\iota}$ is the average performance in the internal test set, and $s$ is the standard deviation of the difference between performance obtained in the external test set and that achieved in the internal test set.

## Results

### *Measuring the effect of data leakage in simulated data*

Regarding the imaging site prediction, the results were similar for both CT and FD simulated features. The performances obtained on the leaked internal test set were overly optimistic, i.e., significantly better than those obtained in the external test set, indicating the presence of data leakage. In contrast, the average balanced accuracies recorded on the not leaked test internal set were statistically not different from those of the external test set (except in one case – see details in Table 4).

Moreover, as the number of samples available in each single-center dataset decreases, the effect of data leakage increases (Tables 3 and 4 for CT and FD, respectively). This phenomenon is even more evident in Figure 5, where we reported the difference between the average balanced accuracy obtained in the external test set and that gained in the internal test sets vs. the number of participants in each single-center site for the CT and FD, respectively. When data leakage is present (dashed lines in Figure 5), the difference between the average balanced accuracy in the external test set and that in the internal leaked test set always differs significantly from zero (Bonferroni adjusted p-values $< 10^{-12}$ and $< 10^{-10}$ for CT and FD, respectively) and increases as the number of participants in each single-center dataset decreases. This result has a profound impact because most neuroimaging studies (with *in vivo* data) have single-centers datasets with a number of subjects between 25 and 100. Conversely, when data leakage is not present (solid lines in Figure 5), the difference between the average balanced accuracy in the external test set and that in the internal not leaked test set was approximately zero and remained constant as the number of participants in each single-center dataset changes.

Data leakage was also observed in the age prediction task for both CT and FD features. Similarly to the site prediction task, the performance on the leaked internal test set appears overly optimistic (Tables 5 and 6 for CT and FD, respectively), and the impact of data leakage becomes more pronounced as the number of samples in each single-center dataset decreases (Figure 6).

***Visualization and quantification of the site effect in vivo data***

Quality control of FreeSurfer's outputs resulted in removing 47 subjects based on the overall low quality of cortical reconstruction or segmentation errors in any regions. All brain regions of the remaining 1740 subjects had both CT and FD features. Thus, we have been able to analyze the site effect, the harmonization adjustments, and age prediction on the same subjects for the CT and FD groups of features. The demographic characteristics of the subjects included in the study after the quality control have been reported in Table 7.

The boxplots in Figures 7 and 8 summarize the distribution of the average CT and FD of the cerebral cortex at each imaging site. Specifically, the site effect differs between the two features. For example, in the CHILDHOOD meta-dataset, the ABIDEI-KKI_32ch, ABIDEI-KKI_8ch, and ABIDEII-NYU_1 single-center datasets show the lowest average CT values, while subjects from the ABIDEI-STANFORD dataset have the lowest FD values. Also, for the ADOLESCENCE meta-dataset, the site effect has a different behavior for CT and FD features: for example, ABIDEI-TCD_1 shows the lowest values of CT, while ABIDEI-LEUVEN shows the lowest values of FD. At the same time, in the ADULTHOOD meta-dataset, ABIDEI-SBL has the lowest mean CT values, whereas ABIDEII-BNI_1 has the lowest FD values.

The same result was measured quantitatively using ANCOVA analysis. Indeed, all CT and FD features were significantly different across the single-center datasets (Table 8), but the site effect, measured by the partial $\eta^2$ was different in the two feature sets. In the CHILDHOOD meta-dataset, for example, each cortical region showed a higher partial $\eta^2$ for FD than for CT, suggesting that, in childhood, acquisition characteristics impact more on the structural complexity measure, i.e., FD, than on the cortical thickness. On the other hand, in the ADOLESCENCE meta-dataset, the frontal and temporal lobes (bilaterally), along with the entire structure, show lower partial $\eta^2$ for FD than for CT, whereas the parietal and occipital lobes (bilaterally) have higher partial $\eta^2$ for FD than for CT. Finally, in the ADULTHOOD meta-dataset, only the occipital and temporal lobes (bilaterally) have lower partial $\eta^2$ for FD than CT.

### *Harmonization efficacy*

To assess whether most of the variation in the data was still associated with the site after harmonization, we predicted the imaging site using neuroimaging features grouped by feature type (i.e., CT and FD). Figures 9 and 10 report the average confusion matrices (over 100 repetitions) for CT and FD features, respectively. When predicting the site using the raw data, the main diagonal of the confusion matrix is prominent (i.e., the predicted site is usually the actual site) for both feature groups and each meta-dataset (Figures 9 and 10). On the other hand, when the prediction of the site is performed using harmonized data (through *neuroHarmonize* or *harmonizer* transformer), the impact of the main diagonal of the confusion matrix is weak. The confusion matrices show a vertical pattern indicating that the predicted site is often the same site, regardless of the actual site (Figures 9 and 10). Moreover, the confusion matrix obtained using the *harmonizer* within the machine learning pipeline seems similar to that obtained by harmonizing all the data with *neuroHarmonize* before imaging site prediction. This result suggests that the action of the *harmonizer* resembles that of *neuroHarmonize*, although the model is built on training data only and then applied to test data. The confusion matrices for CT and FD features in the LIFESPAN meta-dataset have also been shown in Figure 11.

Table 9 reports the median balanced accuracies (over 100 repetitions) of imaging site prediction, and the efficacy of the harmonization is shown in Table 10. Specifically, we have reported the pair (age-group permutation test p-value, one-sided Wilcoxon signed-rank test p-value) to statistically assess the removal or reduction of the site effect, respectively. As expected, the median balanced accuracy of site prediction using the raw data was significantly different from the chance level (age-group permutation test p-value > 0.05 for all data), and thus, an actual imaging site effect was present on raw data. After harmonization, with *neuroHarmonize* or *harmonizer* transformer, the site effect was removed (age-group permutation test p-value ≥ 0.05 in Table 10) or only reduced (age-group permutation test p-value < 0.05, but with median balanced accuracy reduced on harmonized data, as statistically measured by the one-sided Wilcoxon signed-rank test p-value < 0.05 in Table 10). Specifically, by performing harmonization using *neuroHarmonize* on all data, we observe that the site effect removal seems to be ensured in all analyses performed except for the imaging site predictions using FD features in the ADOLESCENCE and ADULTHOOD meta-datasets (age-group permutation test p-value equal to 0.0188 and 0.0002, respectively, in Table 10). We found the same behavior when predicting the imaging site using CT and FD features in the LIFESPAN meta-dataset (age-group permutation test p-value equal to 0.0002 in Table 10). In the latter cases, although significantly different from a random prediction, the balanced accuracies were significantly lower than those obtained using the original data (one-sided Wilcoxon signed-rank test p-values < 0.001 in Table 10), and this indicates a site effect reduction. When applying the *harmonizer* transformer to the data (within the CV), we observed the actual efficacy of the harmonization, without introducing data leakage, as in the previous case. Indeed, we confirmed a complete removal of site effect only in imaging site prediction using CT features in ADULTHOOD meta-dataset (age-group permutation test p-value equal to 0.1064 in Table 10). In all the other cases, the imaging site prediction was significantly different from the chance level (age-group permutation test p-values < 0.05 in Table 10), but the balanced accuracies were significantly lower than those obtained using the original data (one-sided Wilcoxon signed-rank test p-values < 0.001 in Table 10). Thus, the site effect removal measured using data harmonized before the splitting into training and test sets was a clear sign of data leakage even in *in vivo* data.

***Age prediction***

Table 11 reports the median MAE values (over 100 repetitions) of the age prediction model. Overall, MAE values of age prediction using data harmonized with *neuroHarmonize* before the splitting into training and test sets are significantly lower than those obtained using data harmonized with the *harmonizer* within the CV (one-sided Wilcoxon signed-rank p-values < 0.001 for all the cases, except for CT features in the CHILDHOOD meta-dataset, see Table 11). In line with the results of simulations, the data leakage introduced by harmonizing the data all at once leads to an overly optimistic performance.

Finally, in Figures 12 and 13, we reported the age-dependent trends of the average CT and FD of the cerebral cortex without harmonization and harmonized with the *harmonizer* transformer, respectively. In line with previous literature concerning features such as CT and volumes[2,5], also in this study, the harmonized average CT and FD values showed less variability than that observed on raw data.

## Discussion

In this study, we introduced the *harmonizer* transformer, which encapsulates the data harmonization procedure among the preprocessing steps of a machine learning pipeline to avoid data leakage. To this end, we explored the *ComBat* harmonization of CT and FD features extracted from brain $T_1$-weighted MRI data of 1740 healthy subjects aged 5 – 87 years acquired at 36 sites and simulated data. We measured the efficacy of the harmonization process in reducing or removing the unwanted site effect through a two-step assessment comparing the performance in imaging site prediction using harmonized data with that of 1) a random prediction and 2) a prediction using non-harmonized data. Finally, we confirmed how data leakage related to harmonization performed before data splitting leads to overestimating performance in simulated and *in vivo* data.

Using simulated data, we showed that the data leakage effect introduced by performing the harmonization before data splitting is clearly evident and worse when the single-center dataset size is small and comparable with the size of the most common neuroimaging *in vivo* studies. In these simulated experiments, we paid particular attention to comparing different harmonization and machine learning approaches in the same conditions, i.e., the same data splits and using the same number of subjects for harmonization (for this reason, we adopted 80% of the data set size for fitting the *neuroHarmonize* model; indeed using the *harmonizer* approach, the harmonization was computed in the training fold of a 5-fold CV, i.e., using 80% of the samples).

We chose the *ComBat* harmonization method due to its widespread use in the scientific community[7,12,16–34] and its implementation in the *neuroHarmonize* package, which enables the specification of covariates with generic non-linear effects[2]. The efficacy of *ComBat* and its variants has been evaluated by comparing their performance with other harmonization techniques[3,5,6] and by simulating site effects using single-center data [2]. However, various harmonization techniques can be used for features extracted from MRI images. One such method is the residuals harmonization, which employs a global scaling procedure to account for the influence of each site using a pair of parameters (offset and scale). These parameters can be estimated through a linear regression model or a more sophisticated approach that considers non-linearities[5]. Global scaling was initially introduced to harmonize images directly[6]. The adjusted residuals harmonization, an advancement of the residuals harmonization, integrates biological covariates (such as age, sex, and diseases) into the linear regression model, facilitating the removal of unwanted site effects while maintaining biological variability[5]. Lastly, the Correcting Covariance Batch Effects (*CovBat*) method is a recent variant of the *ComBat* method that aims to address site effects in the mean, variance, and covariance of the neuroimaging features[84].

It is important to note that this study was the first in which the efficacy of the harmonization procedure of neuroimaging data has been evaluated by comparing the accuracy of the imaging site prediction also to the chance level. Indeed, previous works have consistently shown a decrease in the accuracy of the imaging site prediction after harmonization, but without applying a significance test, and thus it was not known whether the site effect was removed or only reduced [see, e.g., Ref.[2] and Ref.[5]]. As hypothesized, there was a real imaging site effect on the raw data (age-group permutation test p-value < 0.05 for all data). The site effect was either eliminated or only reduced after data harmonization with *neuroHarmonize* or *harmonizer* transformer. Specifically, the difference between the efficacy of harmonization by applying *neuroHarmonize* on all data or *harmonizer* within the CV was expected because, in the former case, data leakage is present leading to a falsely overestimated performance, i.e., an age-group permutation test p-value ≥ 0.05 and a lower median balanced accuracy (Tables 7 and 8). On the one hand, the complete removal of the imaging site measured using the data harmonized with *neuroHarmonize* was only apparent. Indeed, using the *harmonizer* within the CV, the imaging site effect was completely removed only for CT features in the ADULTHOOD meta-dataset. In line with the results of the simulations, we noted that the median balanced accuracies obtained by performing site prediction with harmonized data using the *neuroHarmonize* show significantly lower values than those observed using the *harmonizer* transformer within the CV (one-sided Wilcoxon signed-rank p-values < 0.001 for all the analyses). The differences found in the median balanced accuracy of imaging site prediction using the *harmonizer* transformer and *neuroHarmonize* emphasize the importance of introducing the *harmonizer* transformer into a machine learning pipeline to avoid data leakage, a source of bias in prediction results. Notably, the procedure used to measure data leakage on the simulated data (i.e., comparing the performance of imaging site prediction between the internal test set of the CV and external test set) was not viable for the *in vivo* data due to the limited sample size in several centers (less than 20 subjects).

Looking at the age-group permutation test p-values for imaging site prediction using data harmonized with *neuroHarmonize* (which were harmonized before splitting into training and test sets), it can be observed that the efficacy of harmonization worsened as the overlap of the age distributions in multicenter meta-datasets decreased (Table 10). Specifically, for CT features, the age-group permutation test p-value was 0.5023 in the CHILDHOOD meta-dataset, which exhibits a good overlap of age distributions (BC = 0.71), but dropped to 0.0002 in the LIFESPAN meta-dataset, which exhibits a BC = 0. Similar behavior was observed for FD features. These results on *in vivo* data are in line with the simulations performed by Pomponio and colleagues[2], which suggested that age-disjoint studies should be challenging to harmonize in the presence of nonlinear age effects[2]. The efficacy of the harmonization performed in CV using the *harmonizer* transformer does not appear seemingly to have a close link to the degree of overlap of the age distributions in the multicenter meta-datasets. This may be explained by the fact that the *harmonizer* transformer handles training data only – randomly chosen within the whole meta-dataset – in the different folds of the CV, and the actual BC values may vary.

The goodness of age prediction using the data harmonized with *neuroHarmonize* before the splitting into training and test sets is falsely increased compared with the use of data harmonized with the *harmonizer* within the CV. Indeed, the median MAE values obtained in predicting age using data harmonized with *neuroHarmonize* before splitting into training and test sets were significantly lower than those estimated using data harmonized with *harmonizer* within the CV (one-sided Wilcoxon signed-rank p-values < 0.001 for all the cases, except for CT features in the CHILDHOOD meta-dataset, see Table 10). These results confirm how data leakage related to data harmonization before splitting them into training and test sets leads to performance overestimation even for *in vivo* data and underlines the importance of encapsulating the data harmonization procedure among the preprocessing steps of a machine learning pipeline.

In previous single-centers studies, we observed that the computation of the FD using the box-counting algorithm with the automated selection of the optimal fractal scaling window implemented in *fractalbrain* best predicted chronological age in two datasets of healthy children and adults among various FD approaches, and more conventional features, such as CT, and gyrification index[59]. In this large multicenter study, we confirmed the more remarkable ability of the FD of the cerebral cortex to predict individual age better than the average CT. In the LIFESPAN meta-dataset, for example, the error in age prediction using CT features (MAE = 7.55 years) was reduced by more than 25% using FD features (MAE = 5.60 years) in line with previous literature[59,68]. This result furtherly confirms that FD conveys additional information to that provided by other conventional structural features[58,59,67,68,86–99].

This study has some limitations. Firstly, to show the utility of encapsulating the data harmonization procedure among the preprocessing steps of a machine learning pipeline to avoid data leakage, we used only the *ComBat* harmonization method. However, other harmonization techniques are available and could be similarly effective, including the recent *CovBat* model, which adds harmonization of covariance between sites[84]. Future research may consider comparing and contrasting the performance of different harmonization methods to identify the optimal approach for specific research questions and data sets.

Secondly, we showed and measured the data leakage effect using simulated and *in vivo* data of CT and FD of the cerebral cortex only. Various other morphological and functional MRI-derived features might be considered. However, the focus of the study was mainly to measure the efficacy of the harmonization and show a possible detrimental effect of data harmonization on the entire dataset before machine learning analysis, and this effect is not relative to the features considered.

Lastly, for site/age prediction, we adopted an XGBoost decision tree with default parameters. It is well known that classification/regression performances may be affected by the value of the hyperparameters, and proper hyperparameter optimization, e.g., through a nested CV, could be adopted. However, this procedure was not feasible in our study because of the relatively small size of data in many centers – an undesired but common scenario in many publicly available datasets. Thus, though this choice was arbitrary, we feel that using the same hyperparameters for both *neuroHarmonize* and *Harmonize* transformer data was reasonable.

In conclusion, we showed that introducing the *harmonizer* transformer, which encapsulates the harmonization procedure among the preprocessing steps of a machine learning pipeline, avoided data leakage. Using *in vivo* data, after *Combat* harmonization, the site effect was completely removed or reduced while preserving the biological variability. We, therefore, suggest that future multicenter imaging studies will include the data harmonization method in the machine learning pipelines and measure the efficacy of the harmonization process.

## Data Availability

The brain MR $T_1$-weighted images that support the findings of this study are available from the following online repositories:

- Autism Brain Imaging Data Exchange (ABIDE): http://fcon_1000.projects.nitrc.org/indi/abide/
- Information eXtraction from Images (IXI) study: https://brain-development.org/ixi-dataset/
- 1000 Functional Connectomes Project (FCP) – ICBM dataset: http://fcon_1000.projects.nitrc.org/fcpClassic/FcpTable.html
- Consortium for Reliability and Reproducibility (CoRR) - *NKI 2 - Nathan Kline Institute (Milham)*: http://fcon_1000.projects.nitrc.org/indi/CoRR/html/index.html

The CT and FD features, derived from brain MR $T_1$-weighted of 1740 subjects, that support the findings of this study, are freely available on Zenodo[100,101]. The simulated CT and FD features that support the findings of this study are freely available on a Zenodo repository[102].

## Code Availability

The source code of the efficacy measurement and *harmonizer* transformer is publicly available in a GitHub repository at https://github.com/Imaging-AI-for-Health-virtual-lab/harmonizer. The following are the versions of software and Python libraries used to obtain the results presented in this study:

- FreeSurfer version 7.1.1. For $T_1$-weighted images belonging to ICBM and NKI2 datasets, we used FreeSurfer version 5.3. ABIDEI $T_1$-weighted images were already processed using FreeSurfer version 5.1.
- fractalbrain toolkit version 1.0
- neuroHarmonize v. 2.1.0 package
- eXtreme Gradient Boosting (XGBoost) version 0.90.

**Acknowledgments**

We wish to thank Federica Giorgini and Riccardo Benedetti for data management, Stefano Orsolini for the technical support in the quality control assessment of all FreeSurfer outputs, and Martina Franco for her preliminary analysis of a part of this data.

**Authors contributions**

**Chiara Marzi**: Conceptualization, Methodology, Software, Validation, Formal analysis, Data curation, Writing – Original draft, Writing – Review & Editing, Visualization

**Marco Giannelli, Andrea Barucci, Carlo Tessa and Mario Mascalchi**: Writing – Review & Editing

**Stefano Diciotti**: Conceptualization, Methodology, Resources, Writing – Original draft, Writing – Review & Editing, Visualization, Supervision, Project administration

**Competing interests**

The authors declare no competing interests.

## Figures

**Figure 1. Age distributions.** Age distributions of participants for CHILDHOOD, ADOLESCENCE, ADULTHOOD, and LIFESPAN meta-datasets, grouped by single-center dataset and sorted by median age.

**Figure 2. 3D box-counting for computation of the FD.** An example of the 3D box-counting algorithm that uses an automated selection of the fractal scaling window through the *fractalbrain* toolkit[59]. *N(s)* is the average number of 3D cubes of side *s* needed to fully enclose the brain structure computed using 20 uniformly distributed random offsets to the grid origin. The regression line within the optimal fractal scaling window, whose slope (sign changed) is the FD, is depicted in red.

**Figure 3. Machine learning pipeline**. A pipeline represents the entire data workflow, combining all transformation steps and machine learning model training. It is essential to automate an end-to-end training/test process without any form of data leakage and improve reproducibility, ease of deployment, and code reuse, especially when complex validation schemes are needed.

**Figure 4. Overview of the analysis of simulated data for each experiment**. After an external hold-out, we computed the performance of a site prediction classifier trained using a) the *harmonizer* transformer within the machine learning pipeline (internal not leaked test set) and b) harmonizing all data with *neuroHarmonize* before imaging site prediction (internal leaked test set). Secondly, we compared these performances with that observed on an external test set never used for harmonization and training.

**Figure 5. Imaging site prediction results with CT and FD simulated data**. We reported the difference between the average balanced accuracy obtained in the external test set and that gained in the internal test sets (dotted line for leaked internal test set and solid line for not leaked internal test set) and Cohen's d effect size vs. the number of participants per single-center dataset n. The cross marker indicates a significant difference between balanced accuracy distributions (one-tailed paired t-test Bonferroni adjusted p-value $< 10^{-9}$ and $< 10^{-10}$ for CT and FD, respectively). The colors and line types in Cohen's d plots are consistent with those employed in the other plots.

**Figure 6. Age prediction results with CT and FD simulated data.** We reported the difference between the average MAE obtained in the external test set and that gained in the internal test sets (dotted line for leaked internal test set and solid line for not leaked internal test set) and Cohen's d effect size vs. the number of participants per single-center dataset n. The cross marker indicates a significant difference between balanced accuracy distributions (see Tables 5 and 6 for details). The colors and line types in Cohen's d plots are consistent with those employed in the other plots.

**Figure 7. Boxplot of the average CT of the cerebral cortex**. The boxplots of the average CT and FD of the cerebral cortex without harmonization are shown for the CHILDHOOD, ADOLESCENCE, ADULTHOOD, and LIFESPAN meta-datasets.

**Figure 8. Boxplot of the average FD of the cerebral cortex**. The boxplots of the average CT and FD of the cerebral cortex without harmonization are shown for the CHILDHOOD, ADOLESCENCE, ADULTHOOD, and LIFESPAN meta-datasets.

**Figure 9. Confusion matrices of site prediction using CT features.** Each confusion matrix was normalized for the number of subjects belonging to each site. In this way, the sum of the matrix cells of each row gives 1. The confusion matrix obtained using the *harmonizer* within the machine learning pipeline seems similar to that obtained by harmonizing all the data with *neuroHarmonize* before imaging site prediction, even though the model is built on training data only and then applied to test data.

**Figure 10. Confusion matrices of site prediction using FD features.** Each confusion matrix was normalized for the number of subjects belonging to each site. In this way, the sum of the matrix cells of each row gives 1. The confusion matrix obtained using the *harmonizer* within the machine learning pipeline seems similar to that obtained by harmonizing all the data with *neuroHarmonize* before imaging site prediction, even though the model is built on training data only and then applied to test data.

**Figure 11. Confusion matrices of site prediction using CT and FD features in the LIFESPAN meta-dataset.** Each confusion matrix was normalized for the number of subjects belonging to each site. In this way, the sum of the matrix cells of each row gives 1. The confusion matrix obtained using the *harmonizer* within the machine learning pipeline seems

similar to that obtained by harmonizing all the data with *neuroHarmonize* before imaging site prediction, even though the model is built on training data only and then applied to test data.

**Figure 12. Scatterplot of the average CT of the cerebral cortex vs. age.** The plot of the average CT of the cerebral cortex vs. age is shown for the CHILDHOOD, ADOLESCENCE, ADULTHOOD, and LIFESPAN meta-datasets without and with harmonization using the *harmonizer* transformer. In the latter case, we considered only the first CV among the 100 repetitions. Specifically, for each subject, we plotted the harmonized value obtained in the fold when the subject was included in the test set.

**Figure 13. Scatterplot of the FD of the cerebral cortex vs. age**. The plot of the FD of the cerebral cortex vs. age is shown for the CHILDHOOD, ADOLESCENCE, ADULTHOOD, and LIFESPAN meta-datasets without and with harmonization using the *harmonizer* transformer. In the latter case, we considered only the first CV among the 100 repetitions. Specifically, for each subject, we plotted the harmonized value obtained in the fold when the subject was included in the test set.

# Tables

## Table 1. Scanning parameters for each single-center dataset.

| Dataset | Manufacturer and model | Magnetostatic field (T) | In-plane resolution (mm × mm) | Slice thickness (mm) | TR (ms) | TE (ms) | TI (ms) | FA (°) |
|---|---|---|---|---|---|---|---|---|
| ABIDEI-CALTECH | Siemens MAGNETOM Trio | 3 | 1.0 × 1.0 | 1.0 | 1590 | 2.73 | 800 | 10 |
| ABIDEI-CMU | Siemens MAGNETOM Verio | 3 | 1.0 × 1.0 | 1.0 | 1870 | 2.48 | 1100 | 8 |
| ABIDEI-KKI | Philips Achieva | 3 | 1.0 × 1.0 | 1.0 | 8.0 | 3.7 | 843 | 8 |
| ABIDEI-LEUVEN | Philips Intera | 3 | 0.98 × 0.98 | 1.20 | 9.6 | 4.6 | 885 | 8 |
| ABIDEI-MAX_MUN | Siemens MAGNETOM Verio | 3 | 1.0 × 1.0 | 1.0 | 1800 | 3.06 | 900 | 9 |
| ABIDEI-NYU | Siemens MAGNETOM Allegra | 3 | 1.3 × 1.0 | 1.3 | 2530 | 3.25 | 1100 | 7 |
| ABIDEI-OHSU | Siemens MAGNETOM Trio | 3 | 1.0 × 1.0 | 1.1 | 2300 | 3.58 | 900 | 10 |
| ABIDEI-OLIN | Siemens MAGNETOM Allegra | 3 | 1.0 × 1.0 | 1.0 | 2500 | 2.74 | 900 | 8 |
| ABIDEI-PITT | Siemens MAGNETOM Allegra | 3 | 1.1 × 1.1 | 1.1 | 2100 | 3.93 | 1000 | 7 |
| ABIDEI-SBL | Philips Intera | 3 | 1.0 × 1.0 | 1.0 | 9.0 | 3.5 | 144 | 8 |
| ABIDEI-SDSU | GE Discovery MR750 | 3 | 1.0 × 1.0 | 1.0 | 11.08 | 4.3 | 600 | 45 |
| ABIDEI-STANFORD | GE Signa | 3 | 0.859 × 1.500 | 0.859 | 8.4 | 1.8 | NA | 15 |
| ABIDEI-TRINITY | Philips Achieva | 3 | 1.0 × 1.0 | 1.0 | 8.5 | 3.9 | 1060 | 8 |
| ABIDEI-UCLA | Siemens MAGNETOM Trio | 3 | 1.0 × 1.0 | 1.2 | 2300 | 2.84 | 853 | 9 |
| ABIDEI-UM | GE Signa | 3 | NA | 1.2 | NA | 1.8 | NA | 15 |
| ABIDEI-USM | Siemens MAGNETOM Trio | 3 | 1.0 × 1.0 | 1.2 | 2300 | 2.91 | 900 | 9 |
| ABIDEII-BNI_1 | Philips Ingenia | 3 | 1.1 × 1.1 | 1.2 | 6.7 | 3.1 | 799 | 9 |
| ABIDEII-EMC_1 | GE Discovery MR750 | 3 | 1.1 × 1.1 | 1.2 | 6.7 | 3.1 | 350 | 9 |
| ABIDEII-ETH_1 | Philips Achieva | 3 | 0.9 × 0.9 | 0.9 | 8.4 | 3.9 | 1150 | 8 |
| ABIDEII-GU_1 | Siemens MAGNETOM Trio | 3 | 1.0 × 1.0 | 1.0 | 2530 | 3.5 | 1100 | 7 |
| ABIDEII-IP_1 | Philips Achieva | 1.5 | 1.0 × 1.0 | 1.0 | 25 | 5.6 | NA | 30 |
| ABIDEII-IU_1 | Siemens TrioTim | 3 | 0.7 × 0.7 | 0.7 | 2400 | 2.3 | 1000 | 8 |
| ABIDEII-KKI_32ch | Philips Achieva | 3 | 0.95 × 0.96 | 1.0 | 8.2 | 3.7 | 753 | 8 |
| ABIDEII-KKI_8ch | Philips Achieva | 3 | 1.0 × 1.0 | 1.0 | 8.0 | 3.7 | 843 | 8 |
| ABIDEII-NYU_1 | Siemens Allegra | 3 | 1.3 × 1.0 | 1.33 | 2530 | 3.25 | 1100 | 7 |
| ABIDEII-OHSU_1 | Siemens TrioTim | 3 | 1.0 × 1.0 | 1.1 | 2300 | 3.58 | 900 | 10 |
| ABIDEII-SDSU_1 | GE Discovery MR750 | 3 | 1.0 × 1.0 | 1.0 | 8.136 | 3.172 | 600 | 8 |
| ABIDEII-TCD_1 | Philips Intera Achieva | 3 | 0.9 × 0.9 | 0.9 | 8.4 | 3.9 | 1150 | 8 |
| ABIDEII-UCD_1 | Siemens TrioTim | 3 | 1.0 × 1.0 | 1.0 | 2000 | 3.16 | 1050 | 8 |
| ABIDEII-UCLA_1 | Siemens TrioTim | 3 | 1.0 × 1.0 | 1.2 | 2300 | 2.86 | 853 | 9 |
| ABIDEII-USM_1 | Siemens TrioTim | 3 | 1.0 × 1.0 | 1.2 | 2300 | 2.91 | 900 | 9 |
| ICBM | NA | 3 | 1.0 × 1.0 | 1.0 | NA | NA | NA | NA |
| IXI-Guys | Philips Gyroscan Intera | 1.5 | NA | NA | 9.813 | 4.603 | NA | 8 |
| IXI-HH | Philips Intera | 3 | NA | NA | 9.6 | 4.6 | NA | 8 |
| IXI-IOP | NA | NA | NA | NA | NA | NA | NA | NA |
| NKI2 | NA | 3 | 1.0 × 1.0 | 1.0 | NA | NA | NA | NA |

FA, flip angle; NA, not available; TE, echo time; TI, inversion time; TR, repetition time.

**Table 2.** Description of the demographic characteristics of each meta-dataset.

| **Meta-dataset** | **# of single-center datasets included** | **# participants** | **Females (%)** | **Age range min-max in years** | **Age median (IQR) in years** | **Age distributions BC** |
|---|---|---|---|---|---|---|
| CHILDHOOD | 11 | 442 | 34.39 | 5.89 – 13.0 | 9.91 (2.08) | 0.71 |
| ADOLESCENCE | 9 | 222 | 15.32 | 11.0 – 20.0 | 14.22 (3.3) | 0.45 |
| ADULTHOOD | 12 | 814 | 47.42 | 18.0 – 86.32 | 42.5 (30.74) | 0 |
| LIFESPAN | 36 | 1787 | 35.65 | 5.89 – 86.32 | 19.98 (27.94) | 0 |

BC, Bhattacharyya coefficient, IQR : interquartile range.

**Table 3.** Imaging site prediction results with CT simulated data. Average balanced accuracy (standard deviation) obtained in the external and internal test sets.

| ***k*** | ***n*** | **External test set** | **Leaked internal test set** | **Not leaked internal test set** |
|---|---|---|---|---|
| 3 | 25 | 0.330 (0.080) | 0.253 (0.050)* | 0.340 (0.049) |
| 3 | 50 | 0.330 (0.052) | 0.269 (0.035)* | 0.335 (0.034) |
| 3 | 100 | 0.347 (0.037) | 0.299 (0.033)* | 0.347 (0.032) |
| 3 | 250 | 0.321 (0.023) | 0.288 (0.016)* | 0.320 (0.015) |
| 10 | 25 | 0.096 (0.026) | 0.054 (0.016)* | 0.100 (0.020) |
| 10 | 50 | 0.098 (0.019) | 0.072 (0.012)* | 0.108 (0.013) |
| 10 | 100 | 0.096 (0.013) | 0.071 (0.007)* | 0.097 (0.007) |
| 10 | 250 | 0.100 (0.008) | 0.085 (0.005)* | 0.102 (0.005) |
| 36 | 25 | 0.025 (0.007) | 0.010 (0.002)* | 0.027 (0.004) |
| 36 | 50 | 0.027 (0.005) | 0.014 (0.002)* | 0.027 (0.002) |
| 36 | 100 | 0.028 (0.004) | 0.018 (0.001)* | 0.029 (0.002) |
| 36 | 250 | 0.027 (0.002) | 0.021 (0.001)* | 0.028 (0.001) |

*k* is the number of single-center datasets, each one containing *n* participants. CT: cortical thickness.

*One-tailed paired t-test Bonferroni adjusted p-value $< 10^{-12}$ for the comparison with the external test set average balanced accuracy.

**Table 4.** Imaging site prediction results with FD simulated data. Average balanced accuracy (standard deviation) obtained in the external and internal test sets.

| ***k*** | ***n*** | **External test set** | **Leaked internal test set** | **Not leaked internal test set** |
|---|---|---|---|---|
| 3 | 25 | 0.358 (0.078) | 0.288 (0.048)* | 0.369 (0.046) |
| 3 | 50 | 0.324 (0.060) | 0.270 (0.045)* | 0.333 (0.045) |
| 3 | 100 | 0.352 (0.047) | 0.296 (0.029)* | 0.347 (0.027) |
| 3 | 250 | 0.328 (0.021) | 0.292 (0.019)* | 0.325 (0.017) |
| 10 | 25 | 0.084 (0.026) | 0.049 (0.014)* | 0.089 (0.017) |
| 10 | 50 | 0.104 (0.017) | 0.068 (0.011)* | 0.105 (0.013) |
| 10 | 100 | 0.098 (0.013) | 0.072 (0.008)* | 0.097 (0.009) |
| 10 | 250 | 0.097 (0.008) | 0.084 (0.005)* | 0.100 (0.005) |
| 36 | 25 | 0.024 (0.007) | 0.010 (0.002)* | 0.026 (0.004) |
| 36 | 50 | 0.028 (0.005) | 0.013 (0.002)* | 0.026 (0.002)^ |
| 36 | 100 | 0.028 (0.003) | 0.017 (0.002)* | 0.028 (0.002) |
| 36 | 250 | 0.028 (0.003) | 0.021 (0.001)* | 0.028 (0.001) |

*k* is the number of single-center datasets, each one containing *n* participants. FD: fractal dimension.

*One-tailed paired t-test Bonferroni adjusted p-value < $10^{-10}$ for the comparison with the external test set average balanced accuracy.

^One-tailed t-test Bonferroni adjusted p-value = 0.003 for the comparison with the external test set average balanced accuracy.

**Table 5.** Age prediction results with CT simulated data. Average MAE (standard deviation) obtained in the external and internal test sets.

| ***k*** | ***n*** | **External test set** | **Leaked internal test set** | **Not leaked internal test set** |
|---|---|---|---|---|
| 3 | 25 | 6.908 (1.199) | 6.561 (0.676) | 6.997 (0.621) |
| 3 | 50 | 5.409 (0.524) | 5.076 (0.322)* | 5.279 (0.301) |
| 3 | 100 | 5.326 (0.314) | 5.273 (0.207) | 5.376 (0.196) |
| 3 | 250 | 4.596 (0.144) | 4.643 (0.137) | 4.670 (0.132) |
| 10 | 25 | 5.473 (0.402) | 5.099 (0.270)* | 5.651 (0.273) |
| 10 | 50 | 4.930 (0.200) | 4.891 (0.166) | 5.128 (0.153) |
| 10 | 100 | 4.823 (0.163) | 4.589 (0.116)* | 4.701 (0.109)^ |
| 10 | 250 | 4.529 (0.097) | 4.556 (0.068) | 4.604 (0.068) |
| 36 | 25 | 8.114 (0.308) | 7.272 (0.223)* | 7.931 (0.201)^ |
| 36 | 50 | 7.478 (0.175) | 7.117 (0.150)* | 7.424 (0.146) |
| 36 | 100 | 7.175 (0.129) | 7.147 (0.083)* | 7.300 (0.083) |
| 36 | 250 | 7.118 (0.073) | 7.076 (0.054)* | 7.139 (0.050) |

*k* is the number of single-center datasets, each one containing *n* participants. CT: cortical thickness.

*One-tailed t-test Bonferroni adjusted p-value < $10^{-4}$ for the comparison with the external test set MAE.

^One-tailed t-test Bonferroni adjusted p-value < $10^{-6}$ for the comparison with the external test set MAE.

**Table 6.** Age prediction results with FD simulated data. Average MAE (standard deviation) obtained in the external and internal test sets.

| *k* | *n* | External test set | Leaked internal test set | Not leaked internal test set |
|---|---|---|---|---|
| 3 | 25 | 6.736 (0.857) | 6.060 (0.682)* | 6.526 (0.672) |
| 3 | 50 | 5.280 (0.500) | 4.971 (0.333)* | 5.133 (0.322) |
| 3 | 100 | 5.069 (0.282) | 4.827 (0.225)* | 4.896 (0.220)^ |
| 3 | 250 | 4.423 (0.159) | 4.486 (0.138) | 4.516 (0.134) |
| 10 | 25 | 5.644 (0.462) | 5.128 (0.319)* | 5.691 (0.299) |
| 10 | 50 | 5.263 (0.288) | 5.007 (0.269)* | 5.254 (0.256) |
| 10 | 100 | 4.830 (0.144) | 4.599 (0.134)* | 4.707 (0.124)^ |
| 10 | 250 | 4.383 (0.072) | 4.422 (0.065) | 4.467 (0.062) |
| 36 | 25 | 8.199 (0.298) | 7.653 (0.162)* | 8.281 (0.144) |
| 36 | 50 | 7.608 (0.201) | 7.398 (0.136)* | 7.707 (0.133) |
| 36 | 100 | 7.119 (0.144) | 7.149 (0.084) | 7.307 (0.078) |
| 36 | 250 | 7.174 (0.098) | 7.102 (0.061)* | 7.160 (0.057) |

*k* is the number of single-center datasets, each one containing *n* participants. CT: cortical thickness.

*One-tailed t-test Bonferroni adjusted p-value < $10^{-7}$ for the comparison with the external test set MAE.

^One-tailed t-test Bonferroni adjusted p-value < 0.0001 for the comparison with the external test set MAE.

**Table 7.** Demographic characteristics of the subjects remaining after quality control and who entered into the analyses.

| Site | # participants | Females (%) | Age min – max (years) | Age mean (SD) (years) | Age median (IQR) (years) |
|---|---|---|---|---|---|
| ABIDEI-CALTECH | 19 | 21.05 | 17.0 - 56.2 | 28.87 (11.21) | 23.6 (15.85) |
| ABIDEI-CMU | 13 | 23.08 | 20.0 - 40.0 | 26.85 (5.74) | 27.0 (9.0) |
| ABIDEI-KKI | 32 | 28.12 | 8.07 - 12.77 | 10.15 (1.28) | 9.97 (1.58) |
| ABIDEI-LEUVEN | 35 | 14.29 | 12.2 - 29.0 | 18.17 (4.99) | 16.6 (7.95) |
| ABIDEI-MAX_MUN | 33 | 12.12 | 7.0 - 48.0 | 26.21 (9.8) | 26.0 (9.0) |
| ABIDEI-NYU | 104 | 24.04 | 6.47 - 31.78 | 15.87 (6.25) | 14.4 (8.78) |
| ABIDEI-OHSU | 15 | 0 | 8.2 - 11.99 | 10.06 (1.08) | 10.08 (1.31) |
| ABIDEI-OLIN | 14 | 14.29 | 10.0 - 23.0 | 16.93 (3.63) | 16.5 (5.75) |
| ABIDEI-PITT | 27 | 14.81 | 9.44 - 33.24 | 18.88 (6.64) | 17.13 (8.3) |
| ABIDEI-SBL | 15 | 0 | 20.0 - 42.0 | 33.73 (6.61) | 36.0 (11.5) |
| ABIDEI-SDSU | 21 | 28.57 | 8.67 - 16.88 | 14.18 (1.94) | 14.1 (2.62) |
| ABIDEI-STANFORD | 12 | 33.33 | 7.75 - 12.43 | 10.2 (1.68) | 9.76 (2.9) |
| ABIDEI-TRINITY | 25 | 0 | 12.04 - 25.66 | 17.08 (3.77) | 15.91 (5.25) |
| ABIDEI-UCLA | 42 | 14.29 | 9.21 - 17.79 | 12.92 (1.96) | 12.7 (2.15) |
| ABIDEI-UM | 72 | 25 | 8.2 - 28.8 | 14.85 (3.64) | 14.8 (4.95) |
| ABIDEI-USM | 43 | 0 | 8.77 - 39.39 | 21.36 (7.64) | 19.76 (10.4) |
| ABIDEII-BNI_1 | 29 | 0 | 18.0 - 64.0 | 39.59 (15.09) | 43.0 (27.0) |
| ABIDEII-EMC_1 | 25 | 16 | 6.33 - 10.12 | 8.16 (1.03) | 8.19 (1.52) |
| ABIDEII-ETH_1 | 24 | 0 | 13.83 - 30.67 | 23.88 (4.5) | 24.0 (6.79) |
| ABIDEII-GU_1 | 51 | 49.02 | 8.06 - 13.8 | 10.49 (1.72) | 10.43 (3.04) |
| ABIDEII-IP_1 | 32 | 68.75 | 8.07 - 46.6 | 24.05 (11.64) | 22.49 (14.71) |
| ABIDEII-IU_1 | 20 | 25 | 19.0 - 37.0 | 23.75 (4.9) | 22.0 (4.25) |
| ABIDEII-KKI_32ch | 45 | 26.67 | 8.06 - 12.67 | 10.42 (1.26) | 10.27 (1.67) |
| ABIDEII-KKI_8ch | 107 | 40.19 | 8.02 - 12.9 | 10.3 (1.17) | 10.3 (1.67) |
| ABIDEII-NYU_1 | 30 | 6.67 | 5.89 - 23.81 | 9.52 (3.33) | 9.11 (3.12) |
| ABIDEII-OHSU_1 | 55 | 52.73 | 8.0 - 14.0 | 10.4 (1.64) | 10.0 (2.5) |
| ABIDEII-SDSU_1 | 25 | 8 | 8.1 - 17.7 | 13.25 (3.04) | 13.0 (5.2) |
| ABIDEII-TCD_1 | 21 | 0 | 10.25 - 20.0 | 15.61 (3.12) | 15.25 (5.25) |
| ABIDEII-UCD_1 | 14 | 28.57 | 12.25 - 17.17 | 14.8 (1.71) | 14.75 (2.65) |
| ABIDEII-UCLA_1 | 15 | 33.33 | 7.76 - 14.09 | 9.81 (2.18) | 9.02 (1.93) |
| ABIDEII-USM_1 | 16 | 18.75 | 11.5 - 36.15 | 23.98 (7.8) | 23.78 (10.72) |
| ICBM | 86 | 52.33 | 19.0 - 85.0 | 44.19 (17.92) | 44.5 (31.5) |
| IXI-Guys | 309 | 55.99 | 20.07 - 86.2 | 50.75 (15.83) | 53.41 (25.59) |
| IXI-HH | 178 | 52.25 | 20.17 - 81.94 | 47.12 (16.65) | 47.38 (29.16) |
| IXI-IOP | 63 | 65.08 | 19.98 - 86.32 | 40.58 (15.41) | 35.46 (17.33) |
| NKI2 | 73 | 41.1 | 6.0 - 17.0 | 11.85 (3.14) | 12.0 (6.0) |

IQR: interquartile range; SD: standard deviation.

**Table 8.** ANCOVA results on raw data. The partial $\eta^2$ values are reported for each region/feature pair analyzed separately in each meta-dataset. All p-values are < 0.001.

| | **CHILDHOOD meta-dataset** | | **ADOLESCENCE meta-dataset** | | **ADULTHOOD meta-dataset** | | **LIFESPAN meta-dataset** | |
|---|---|---|---|---|---|---|---|---|
| **Region** | **CT** | **FD** | **CT** | **FD** | **CT** | **FD** | **CT** | **FD** |
| Entire cortex | 0.42 | 0.52 | 0.34 | 0.32 | 0.28 | 0.31 | 0.33 | 0.41 |
| lh cortex | 0.44 | 0.49 | 0.32 | 0.36 | 0.28 | 0.36 | 0.33 | 0.43 |
| rh cortex | 0.39 | 0.42 | 0.34 | 0.36 | 0.26 | 0.27 | 0.32 | 0.39 |
| lh cortex frontal lobe | 0.34 | 0.43 | 0.48 | 0.42 | 0.22 | 0.26 | 0.35 | 0.35 |
| lh cortex occipital lobe | 0.26 | 0.37 | 0.31 | 0.46 | 0.44 | 0.30 | 0.36 | 0.41 |
| lh cortex temporal lobe | 0.57 | 0.65 | 0.25 | 0.21 | 0.33 | 0.25 | 0.44 | 0.51 |
| lh cortex parietal lobe | 0.26 | 0.39 | 0.25 | 0.39 | 0.21 | 0.27 | 0.24 | 0.37 |
| rh cortex frontal lobe | 0.21 | 0.34 | 0.53 | 0.41 | 0.18 | 0.26 | 0.30 | 0.33 |
| rh cortex occipital lobe | 0.21 | 0.41 | 0.30 | 0.46 | 0.46 | 0.34 | 0.37 | 0.46 |
| rh cortex temporal lobe | 0.59 | 0.66 | 0.23 | 0.17 | 0.35 | 0.29 | 0.48 | 0.54 |
| rh cortex parietal lobe | 0.27 | 0.47 | 0.32 | 0.38 | 0.24 | 0.28 | 0.29 | 0.39 |

CT: cortical thickness; FD: fractal dimension; lh: left hemisphere; rh: right hemisphere.

**Table 9.** Site prediction results. The median and the interquartile range of the balanced accuracy over 100 repetitions of the 5-fold CV have been reported. In bold, we have highlighted significant falsely overestimated performance due to data leakage (the median balanced accuracy in imaging site prediction using data harmonized with *neuroHarmonize* is lower, i.e., better performance, than that estimated using data harmonized with the *harmonizer* transformer within the CV – one-sided Wilcoxon signed-rank test p-values < 0.001 for all the analyses).

| | **Balanced accuracy median (IQR)** | | |
|---|---|---|---|
| **Meta-dataset and feature type** | **Without harmonization** | **Harmonization with *neuroHarmonize*** | **Harmonization with *harmonizer* transformer** |
| CHILDHOOD | | | |
| CT | 0.45 (0.03) | **0.09 (0.01)** | 0.13 (0.02) |
| FD | 0.35 (0.02) | **0.09 (0.01)** | 0.13 (0.02) |
| ADOLESCENCE | | | |
| CT | 0.45 (0.03) | **0.13 (0.02)** | 0.15 (0.03) |
| FD | 0.43 (0.04) | **0.16 (0.03)** | 0.22 (0.04) |
| ADULTHOOD | | | |
| CT | 0.40 (0.03) | **0.09 (0.01)** | 0.09 (0.01) |
| FD | 0.29 (0.02) | **0.12 (0.01)** | 0.13 (0.01) |
| LIFESPAN | | | |
| CT | 0.28 (0.01) | **0.06 (0.01)** | 0.07 (0.01) |
| FD | 0.22 (0.01) | **0.08 (0.01)** | 0.10 (0.01) |

CT: cortical thickness; CV: cross-validation; FD: fractal dimension; IQR: interquartile range.

**Table 10.** Harmonization efficacy. The age-group permutation test p-value and one-sided Wilcoxon signed-rank test p-value have been reported. The permutation test p-value indicates whether the site effect has been removed (i.e., p-value greater than 0.05 means that the imaging site prediction is not different from a random prediction). One-sided Wilcoxon signed-rank test p-value indicates whether the site effect has been reduced (i.e., p-value less than 0.05 means that the prediction of imaging site using the harmonized features obtains a balanced accuracy significantly less than that estimated using raw data).

| | **Harmonization efficacy (age-group permutation test p-value, One-sided Wilcoxon signed-rank test p-value)** | |
|---|---|---|
| **Meta-dataset and feature type** | **Harmonization with *neuroHarmonize*** | **Harmonization with *harmonizer* transformer** |
| CHILDHOOD | | |
| CT | (0.5363, $10^{-18}$) | (0.0036, $10^{-18}$) |
| FD | (0.5853, $10^{-18}$) | (0.0268, $10^{-18}$) |
| ADOLESCENCE | ADOLESCENCE | |
| CT | (0.3559, $10^{-18}$) | (0.0484, $10^{-18}$) |
| FD | (0.0090, $10^{-18}$) | (0.0002, $10^{-18}$) |
| ADULTHOOD | ADULTHOOD | |
| CT | (0.4545, $10^{-18}$) | (0.4727, $10^{-18}$) |
| FD | (0.0042, $10^{-18}$) | (0.0006, $10^{-18}$) |
| LIFESPAN | | |
| CT | (0.1128, $10^{-18}$) | (0.0006, $10^{-18}$) |
| FD | (0.0002, $10^{-18}$) | (0.0002, $10^{-18}$) |

CT: cortical thickness; FD: fractal dimension.

**Table 11.** Age prediction results. The median MAE and the relative standard deviation over 100 repetitions have been reported. In bold, we have highlighted significant falsely overestimated performance due to data leakage (the median MAE in predicting age using data harmonized with *neuroHarmonize* is lower than that estimated using data harmonized with the *harmonizer* transformer within the CV – one-sided Wilcoxon signed-rank test p-values < 0.05 for all the analyses, except for the CT features of the CHILDHOOD meta-dataset).

| | **MAE median (IQR)** | | |
|---|---|---|---|
| **Meta-dataset and feature type** | **Harmonization with *neuroHarmonize*** | **Harmonization with *harmonizer*** | **One-sided Wilcoxon signed-rank test p-value** |
| CHILDHOOD | | | |
| CT | 1.31 (0.03) | 1.28 (0.03) | 1 |
| FD | **1.16 (0.02)** | 1.18 (0.02) | $10^{-13}$ |
| ADOLESCENCE | | | |
| CT | **1.69 (0.06)** | 1.76 (0.07) | $10^{-15}$ |
| FD | **1.56 (0.06)** | 1.59 (0.06) | $10^{-6}$ |
| ADULTHOOD | | | |
| CT | **10.85 (0.12)** | 10.88 (0.14) | 0.007 |
| FD | **8.68 (0.13)** | 8.73 (0.13) | $10^{-5}$ |
| LIFESPAN | | | |
| CT | **7.35 (0.06)** | 7.55 (0.09) | $10^{-18}$ |
| FD | **5.48 (0.04)** | 5.60 (0.07) | $10^{-18}$ |

CT: cortical thickness; CV: cross-validation; FD: fractal dimension; IQR: interquartile range; MAE: median absolute error.

**Supplementary Information**

**Table of contents**

**Table S1.** Description of the demographic characteristics of each single-center dataset. Data from single-center datasets have been included in one specific meta-dataset considering the following age ranges: CHILDHOOD (5 – 13 years), ADOLESCENCE (11 – 20 years), and ADULTHOOD (18 – 87 years). All single-center datasets have also been included in the LIFESPAN meta-dataset.

| Dataset | Institution | # participants | Females (%) | Age range min-max | Age median (IQR) | Meta-dataset |
|---|---|---|---|---|---|---|
| ABIDEI-CALTECH | California Institute of Technology | 19 | 21.05 | 17.0 - 56.2 | 23.6 (15.85) | ADULTHOOD |
| ABIDEI-CMU | Carnegie Mellon University | 13 | 23.08 | 20.0 - 40.0 | 27.0 (9.0) | |
| ABIDEI-KKI | Kennedy Krieger Institute | 33 | 27.27 | 8.07 - 12.77 | 9.97 (1.54) | CHILDHOOD |
| ABIDEI-LEUVEN | University of Leuven | 35 | 14.29 | 12.2 - 29.0 | 16.6 (7.95) | ADOLESCENCE |
| ABIDEI-MAX_MUN | Ludwig Maximilians University Munich | 33 | 12.12 | 7.0 - 48.0 | 26.0 (9.0) | ADULTHOOD |
| ABIDEI-NYU | New York University Langone Medical Center | 105 | 24.76 | 6.47 - 31.78 | 14.38 (8.99) | ADOLESCENCE |
| ABIDEI-OHSU | Oregon Health & Science University | 15 | 0 | 8.2 - 11.99 | 10.08 (1.31) | CHILDHOOD |
| ABIDEI-OLIN | Olin Center, Institute of Living at Hartford Hospital | 16 | 12.5 | 10.0 - 23.0 | 16.5 (6.25) | ADOLESCENCE |
| ABIDEI-PITT | University of Pittsburgh School of Medicine | 27 | 14.81 | 9.44 - 33.24 | 17.13 (8.3) | ADOLESCENCE |
| ABIDEI-SBL | Social Brain Lab BCN NeuroImaging Center, University Medical Center Groningen and Netherlands Institute for Neuroscience | 15 | 0 | 20.0 - 42.0 | 36.0 (11.5) | ADULTHOOD |
| ABIDEI-SDSU | San Diego State University | 22 | 27.27 | 8.67 - 16.88 | 14.42 (2.45) | ADOLESCENCE |
| ABIDEI-STANFORD | Stanford University | 20 | 20 | 7.75 - 12.43 | 9.41 (2.57) | CHILDHOOD |
| ABIDEI-TRINITY | Trinity College Dublin | 25 | 0 | 12.04 - 25.66 | 15.91 (5.25) | ADOLESCENCE |
| ABIDEI-UCLA | University of California, Los Angeles | 47 | 12.77 | 9.21 - 17.79 | 12.68 (2.16) | ADOLESCENCE |
| ABIDEI-UM | University of Michigan | 77 | 23.38 | 8.2 - 28.8 | 14.8 (5.0) | |
| ABIDEI-USM | University of Utah School of Medicine | 43 | 0 | 8.77 - 39.39 | 19.76 (10.4) | |
| ABIDEII-BNI_1 | Barrow Neurological Institute | 29 | 0 | 18.0 - 64.0 | 43.0 (27.0) | ADULTHOOD |
| ABIDEII-EMC_1 | Erasmus University Medical Centre | 27 | 18.52 | 6.33 - 10.12 | 8.19 (1.41) | CHILDHOOD |
| ABIDEII-ETH_1 | ETH Zürich | 24 | 0 | 13.83 - 30.67 | 24.0 (6.79) | ADULTHOOD |
| ABIDEII-GU_1 | Georgetown University | 55 | 49.09 | 8.06 - 13.8 | 10.43 (2.96) | CHILDHOOD |
| ABIDEII-IP_1 | Institut Pasteur and Robert Debré Hospital | 34 | 64.71 | 8.07 - 46.6 | 22.12 (18.8) | ADULTHOOD |
| ABIDEII-IU_1 | Indiana University | 20 | 25 | 19.0 - 37.0 | 22.0 (4.25) | ADULTHOOD |
| ABIDEII-KKI_32ch | Kennedy Krieger Institute | 45 | 26.67 | 8.06 - 12.67 | 10.27 (1.67) | CHILDHOOD |
| ABIDEII-KKI_8ch | Kennedy Krieger Institute | 110 | 40 | 8.02 - 12.9 | 10.3 (1.68) | CHILDHOOD |
| ABIDEII-NYU_1 | New York University Langone Medical Center | 30 | 6.67 | 5.89 - 23.81 | 9.11 (3.12) | CHILDHOOD |
| ABIDEII-OHSU_1 | Oregon Health & Science University | 56 | 51.79 | 8.0 - 14.0 | 10.0 (2.25) | CHILDHOOD |
| ABIDEII-SDSU_1 | San Diego State University | 25 | 8 | 8.1 - 17.7 | 13.0 (5.2) | ADOLESCENCE |
| ABIDEII-TCD_1 | Trinity College Dublin | 21 | 0 | 10.25 - 20.0 | 15.25 (5.25) | ADOLESCENCE |
| ABIDEII-UCD_1 | University of California Davis | 14 | 28.57 | 12.25 - 17.17 | 14.75 (2.65) | |
| ABIDEII-UCLA_1 | University of California, Los Angeles | 16 | 31.25 | 7.76 - 14.09 | 9.01 (1.43) | CHILDHOOD |
| ABIDEII-USM_1 | University of Utah School of Medicine | 16 | 18.75 | 11.5 - 36.15 | 23.78 (10.72) | ADULTHOOD |
| ICBM | International Consortium for Human Brain Mapping | 86 | 52.33 | 19.0 - 85.0 | 44.5 (31.5) | ADULTHOOD |
| IXI-Guys | Guy’s Hospital - London | 313 | 55.59 | 20.07 - 86.2 | 53.41 (26.08) | ADULTHOOD |
| IXI-HH | Hammersmith Hospital - London | 181 | 51.38 | 20.17 - 81.94 | 48.05 (28.25) | ADULTHOOD |
| IXI-IOP | Institute of Psychiatry - London | 67 | 65.67 | 19.98 - 86.32 | 36.16 (19.85) | ADULTHOOD |
| NKI2 | | 73 | 41.1 | 6.0 - 17.0 | 12.0 (6.0) | CHILDHOOD |

IQR: interquartile range